\documentclass[runningheads]{llncs}
\usepackage[utf8]{inputenc}
\usepackage{graphicx}
\usepackage{amsmath,amssymb} 
\usepackage{color}
\usepackage{epsfig}
\usepackage{array}
\usepackage{comment}
\usepackage{algorithm}
\usepackage{algorithmic}
\usepackage{tikz}

\newcommand{\ul}[1]{\underline{#1}}

\definecolor{gold}{HTML}{BD820B}
\definecolor{silver}{HTML}{909090}
\definecolor{bronze}{HTML}{9A5F26}
\newcommand*\circledd[1]{\tikz[baseline=(char.base)]{
            \node[shape=circle,draw,inner sep=0.15pt] (char) {#1};}}           
\newcommand{\first}[1]{%
    {\raisebox{0.8pt}{\footnotesize \color{gold} \circledd{1}}\hspace{3.5pt}{\bf #1}}%
}
\newcommand{\second}[1]{%
    {\raisebox{0.8pt}{\footnotesize \color{silver} \circledd{2}}\hspace{3.5pt}#1}%
}
\newcommand{\third}[1]{%
    {\raisebox{0.8pt}{\footnotesize \color{bronze} \circledd{3}}\hspace{3.5pt}#1}%
}

\graphicspath{{./figures/}}

\begin{document}

\title{FuCoLoT -- A Fully-Correlational Long-Term Tracker
}
\titlerunning{FuCoLoT -- A Fully-Correlational Long-Term Tracker}


\author{
Alan Lukežič\inst{1} \and 
Luka Čehovin Zajc\inst{1} \and
Tomáš Vojíř\inst{2} \and
Jiří Matas\inst{2} \and 
Matej Kristan\inst{1}
}

%

\authorrunning{A. Lukežič et al.} 


\institute{
Faculty of Computer and Information Science, University of Ljubljana, Slovenia \and
Faculty of Electrical Engineering, Czech Technical University in Prague, Czech Republic \\
\email{alan.lukezic@fri.uni-lj.si}
}

\maketitle

\begin{abstract}
We propose FuCoLoT -- a \ul{Fu}lly \ul{Co}rrelational \ul{Lo}ng-term \ul{T}racker. It exploits the novel DCF constrained filter learning method to design a detector that is able to re-detect the target in the whole image efficiently. 
FuCoLoT maintains several correlation filters trained on different time scales that act as the detector components. 
A novel mechanism based on the correlation response is used for tracking failure estimation. FuCoLoT achieves state-of-the-art results on standard short-term benchmarks and it outperforms the current best-performing tracker on the long-term  UAV20L benchmark by over 19\%. 
It has an order of magnitude smaller memory footprint than its best-performing competitors and runs at 15fps in a single CPU thread.
\end{abstract}


\section{Introduction}  \label{sec:introduction}
 
The computer vision community has recently witnessed  significant activity
and  advances of model-free short-term trackers~\cite{otb_pami2015,kristan_vot_tpami2016} which localize a target in a video sequence given a single training example in the first frame.
Current short-term trackers~\cite{danelljan_eccv2016_ccot,Lukezic_CVPR_2017,siamfc_eccv16,Valmadre_2017_CVPR,IGS_TIP2018} localize the target moderately well even in the presence of significant appearance and motion changes and they are robust to short-term occlusions.
Nevertheless, any adaptation at an inaccurate target position leads to gradual corruption of the visual model, drift and irreversible failure. 
Another major source of failures of short-term trackers are significant occlusion and target disappearance from the field of view. These problems are addressed by long-term trackers which combine a short-term tracker with a detector that is capable of reinitializing the tracker.

A long-term tracker development is complex as it entails: (i) the design of the two core components - the short term tracker and the detector, (ii) an algorithm for their interaction including the estimation of tracking and detection uncertainty,  and (iii) the model adaptation strategy. 
Initially, memoryless displacement estimators like the flock-of-trackers~\cite{kalal_pami} and the flow calculated at keypoints~\cite{CMT_CVPR2015} were considered. 
Later, methods applied keypoint detectors~\cite{CMT_CVPR2015,muster_cvpr2015,Pernici2013,Maresca2013}, but these require  large and sufficiently well textured targets. 
Cascades of classifiers~\cite{kalal_pami,LCT_CVPR2015} and more recently deep feature object detection systems~\cite{ptav_iccv2017} have been proposed to deal with diverse targets. 
The drawback is in the significant increase  of computational complexity and the subsequent reduction in the range of possible applications. 
Recent long-term trackers either train the detector on the first frame only~\cite{CMT_CVPR2015,ptav_iccv2017}, thus losing the opportunity to learn target appearance variability or adapt the detector~\cite{LCT_CVPR2015,muster_cvpr2015} and becoming prone to failure due to learning from incorrect training examples.

\begin{figure}[t]
\begin{center}
	\includegraphics[width=0.8\linewidth]{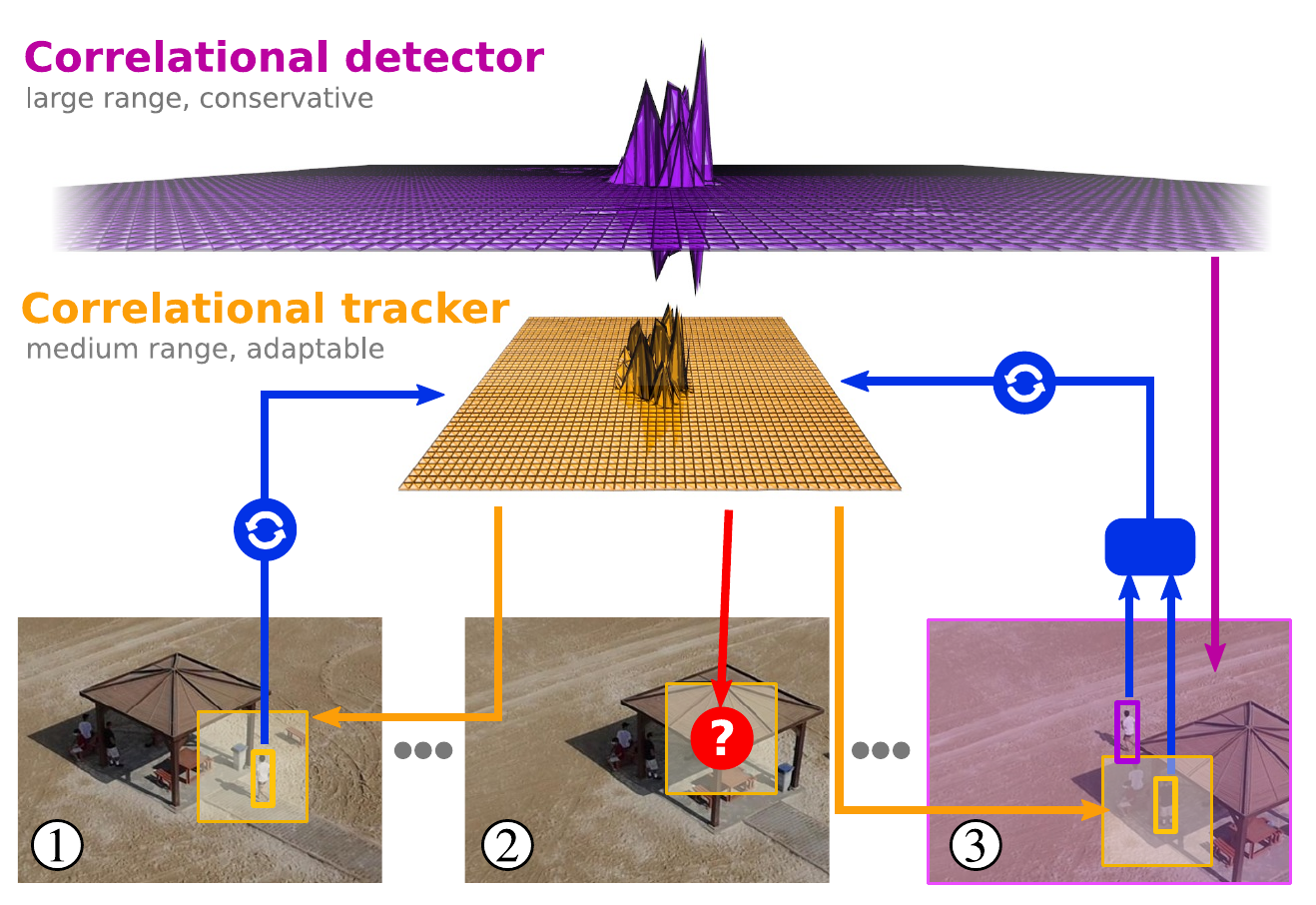}
\end{center}
   \caption{The FuCoLoT tracker framework: A short-term component of FuCoLoT tracks a visible target (1). At occlusion onset (2), the localization uncertainty is detected and the detection correlation filter is activated in parallel to short-term component to account for the two possible hypotheses of uncertain localization. Once the target becomes visible (3), the detector and short-term tracker interact to recover its position. Detector is deactivated once the localization uncertainty drops.}
\label{fig:overview} 
\end{figure}

The paper introduces FuCoLoT - a novel \ul{Fu}lly \ul{Co}rrelational \ul{Lo}ng-term \ul{T}racker.
FuCoLoT is the first long-term tracker that exploits the novel discriminative correlation filter (DCF) learning method based on the ADMM optimization that allows to control the support of the discriminative filter.
The method was first used in CSRDCF~\cite{Lukezic_CVPR_2017} to limit the DCF filter support to the object segmentation and to avoid problems with shapes not well approximated by a rectangle.

The first contribution of the paper is the observation, and its use, that the ADMM optimization allows DCF to search in an area with size unrelated to  the object, e.g. in the whole image. The decoupling of the target and the search region sizes allows implementing the detector as a DCF. Both the short-term tracker and the detector of FuCoLoT are DCFs operating efficiently on the same representation, making FuCoLoT ``fully correlational''.
For some time, DCFs have been the state-of-the-art in short-term tracking, topping a number of recent benchmarks~\cite{otb_pami2015,kristan_vot2016,kristan_vot2015,kristan_vot2016,kristan_vot_tpami2016}.
However, with the standard learning algorithm~\cite{henriques2015tracking}, a correlation filter cannot be used for detection because of two reasons: (i) the dominance of the background in the search regions which necessarily has the same size as the target model and (ii) the effects of the periodic extension on the borders. 
Only recently theoretical breakthroughs~\cite{srdcf_iccv2015,Lukezic_CVPR_2017,cfwlb_cvpr2015} allowed constraining the non-zero filter response to the area covered by the target.

As a second contribution, FuCoLoT uses correlation filters trained on {\it different time scales} as a detector to achieve resistance to occlusions, disappearance, or short-term tracking problems of different duration. Both the detectors and its  short-term tracker is implemented
by a CSRDCF core~\cite{Lukezic_CVPR_2017}, see  Figure~\ref{fig:overview}.
 
The estimation of the relative confidence of the detectors and the short-term tracker, as well as of the localization uncertainty, is facilitated by the fact that both the detector and the short-term tracker output the same representation - the correlation response. We show that this leads to a simple and effective method that controls their interaction.
As another contribution, a stabilizing mechanism is introduced that enables the detector to recover from model contamination.

Extensive experiments show that the proposed FuCoLoT tracker by far outperforms all 
trackers on a long-term benchmark and achieves excellent performance even on short-term benchmarks. FuCoLoT has a small memory footprint, does not require GPUs and runs at 15 fps on a CPU since both the detectors and the short-term tracker enjoy efficient implementation through FFT.

\section{Related work}  \label{sec:related_work}

We briefly overview the most related short-term DCFs and long-term trackers.

\textbf{Short-term DCFs.} Since their inception as the MOSSE tracker~\cite{bolme2010visual}, several advances have made discriminative correlation filters  the most widely used methodology in short-term tracking~\cite{kristan_vot_tpami2016}. 
Major boosts in performance followed introduction of kernels~\cite{henriques2015tracking}, multi-channel formulations~\cite{danelljan2014adaptive,galoogahi_multi_channel_correlation} and scale estimation~\cite{danelljan_dsst_pami,samf_eccv2014}. 
Hand-crafted features have been recently replaced with deep features trained for classification ~\cite{danelljan_eccv2016_ccot,DanelljanCVPR2017} as well as features trained for localization~\cite{Valmadre_2017_CVPR}. 
Another line of research lead to constrained filter learning  approaches~\cite{srdcf_iccv2015,Lukezic_CVPR_2017} that allow learning a filter with the effective size smaller than the training patch.  

\textbf{Long-term trackers.} The long-term trackers combine a short-term tracker with a detector -- an architecture first proposed by Kalal et al.~\cite{kalal_pami} and now commonly used in long-term trackers.
The seminal work of Kalal et al.~\cite{kalal_pami} proposes a memory-less flock of flows as a short-term tracker and a template-based detector run in parallel. 
They propose a P-N learning approach in which the short-term tracker provides training examples for the detector and pruning events are used to reduce contamination of the detector model. The detector is implemented as a cascade to reduce the computational complexity.
 
Another paradigm was pioneered by Pernici et al.~\cite{Pernici2013}. Their approach casts localization as local keypoint descriptors matching with a weak geometrical model. 
They propose an approach to reduce contamination of the keypoints model that occurs at adaptation during occlusion. 
Nebehay et al.~\cite{CMT_CVPR2015} have shown that a keypoint tracker can be utilized even without updating and using pairs of correspondences in a GHT framework to track deformable models. 
Maresca and Petrosino~\cite{Maresca2013} have extend the GHT approach by integrating various descriptors and introducing a conservative updating mechanism. The keypoint methods require a large and well textured target, which limits their application scenarios.

Several methods achieve long-term capabilities by careful model updates and detection of detrimental events like occlusion. 
Grabner et al.~\cite{grabnerECCV2008} proposed an on-line semi-supervised boosting  method that combines a prior and online-adapted classifiers to reduce drifting. 
Chang et al.~\cite{changICIP2011} apply log-polar transform for tracking failure detection. 
Kwak et al.~\cite{kwakICCV2011} proposed occlusion detection by decomposing the target model into a grid of cells and learning an occlusion classifier  for each cell.
Beyer et al.~\cite{Beyer_arxiv2017} proposed a Bayes filter for target loss detection and re-detection for multi-target tracking.

Recent long-term trackers have shifted back to the tracker-detector paradigm of Kalal et al.~\cite{kalal_pami}, mainly due to availability of DCF trackers~\cite{henriques2015tracking} which provide a robust and fast short-term tracking component. 
Ma et al.~\cite{LCT_CVPR2015,LCT_IJCV2018} proposed a combination of KCF tracker~\cite{henriques2015tracking} and a random ferns classifier as a detector. Similarly, Hong et al.~\cite{muster_cvpr2015} combine a KCF tracker with a SIFT-based detector which is also used to detect occlusions.

The most extreme example of using a fast tracker and a slow detector is the recent work of Fan and Ling~\cite{ptav_iccv2017}. 
They combine a DSST~\cite{danelljan_dsst_pami} tracker with a CNN detector~\cite{tao2016sint} which verifies and potentially corrects proposals of the short-term tracker. 
The tracker achieved excellent results on the challenging long-term benchmark~\cite{uav_benchmark_eccv2016}, but requires a GPU, has a huge memory footprint and requires parallel implementation with backtracking to achieve a reasonable runtime.

\section{Fully correlational long-term tracker}  \label{sec:tracking_framework}

In the following we describe the proposed long-term tracking approach based on constrained discriminative correlation filters. The constrained DCF is overviewed in Section~\ref{sec:csrdcf}, Section~\ref{sec:short-term-tracker} overviews the short-term component, Section~\ref{sec:localization_quality} describes detection of tracking uncertainty, Section~\ref{sec:detector} describes the detector and the long-term tracker is described in Section~\ref{sec:tracking_fclt}.

\subsection{Constrained discriminative filter formulation}  \label{sec:csrdcf}
 
FuCoLoT is based on discriminative correlation filters. Given a search region of size $W \times H$ a set of $N_d$ feature channels $\mathbf{f} = \{ \mathbf{f}_d \}_{d=1}^{N_d}$, where $\mathbf{f}_d\in \mathcal{R}^{W\times H}$, are extracted. A set of $N_d$
correlation filters $\mathbf{h} = \{ \mathbf{h}_d \}_{d=1}^{N_d}$, where $\mathbf{h}_d\in \mathcal{R}^{W\times H}$, are correlated with the extracted features and the object position is estimated as the location of the maximum of the weighted correlation responses
\begin{equation}\label{eq:cf_localization}
	\mathbf{r}= \sum\nolimits_{d=1}^{N_d} w_d (\mathbf{f}_d \star \mathbf{h}_d),
\end{equation}
where $\star$ represents circular correlation, which is efficiently implemented by a Fast Fourier Transform and $\{ w_d \}_{d=1}^{N_d}$ are channel weights. The target scale can be efficiently estimated by another correlation filter trained over the scale-space~\cite{danelljan_dsst_pami}.

We apply the recently proposed filter learning technique (CSRDCF~\cite{Lukezic_CVPR_2017}), which uses the alternating direction method of multipliers (ADMM~\cite{admm_boyd2011}) to constrain the learned filter support by a binary segmentation mask.
In the following we provide a brief overview of the learning framework and refer the reader to the original paper~\cite{Lukezic_CVPR_2017} for details. 

\textbf{Constrained learning.} Since feature channels are treated independently, we will assume a single feature channel (i.e., $N_d=1$) in the following. 
A channel feature $\mathbf{f}$ is extracted from a learning region and a fast segmentation method~\cite{kristan_segmentation_accv14} is applied to produce a binary mask $\mathbf{m}\in \{0,1\}^{W \times H}$ that approximately separates the target from the background. 
Next a filter of the same size as the training region is learned, with support constrained by the mask $\mathbf{m}$. The discriminative filter  $\mathbf{h}$  is learned by introducing a dual variable $\mathbf{h}_c$ and minimizing the following augmented Lagrangian
\begin{eqnarray}\label{eq:augmented_lagrange}
	\mathcal{L}(\hat{\mathbf{h}}_c, \mathbf{h}, \hat{\mathbf{l}} | \mathbf{m}) = \| \mathrm{diag}(\hat{\mathbf{f}}) \overline{\hat{\mathbf{h}}}_c  - \hat{\mathbf{g}} \|^2 + \frac{\lambda}{2} \| \mathbf{m} \odot \mathbf{h} \|^2 + \\ \nonumber
2[\hat{\mathbf{l}}^H \mathcal{F}(\mathbf{h}_c - \mathbf{m} \odot \mathbf{h}) ]_\mathrm{Re} + \mu \|  \mathcal{F}(\mathbf{h}_c -  \mathbf{m} \odot \mathbf{h}) \|^2,
\end{eqnarray}
where $\mathbf{g}$ is a desired output, $\hat{\mathbf{l}}$ is a complex Lagrange multiplier, $\hat{(\cdot)}=\mathcal{F}(\cdot)$ denotes Fourier transformed variable, $[\cdot]_\mathrm{Re}$ is an operator that removes the imaginary part and $\mu$ is a non-negative real number.
The solution is obtained via ADMM~\cite{admm_boyd2011} iterations of two closed-form solutions:
\begin{equation}\label{eq:min_hc}
\hat{\mathbf{h}}_c^{i+1} = \big( \hat{\mathbf{f}} \odot \overline{\hat{\mathbf{g}}} + (\mu \mathcal{F}(\mathbf{m} \odot \mathbf{h}^{i}) - \hat{\mathbf{l}}^{i}) \big)  \odot^{-1} \big( \overline{\hat{\mathbf{f}}} \odot \hat{\mathbf{f}} + \mu^{i} \big),
\end{equation}
\begin{equation}\label{eq:min_h}
\mathbf{h}^{i+1} = \mathbf{m} \odot \mathcal{F}^{-1} \big[ \hat{\mathbf{l}}^{i} + \mu^{i} \hat{\mathbf{h}}_c^{i+1} \big] / \big( \frac{\lambda}{2D} + \mu^{i} \big),
\end{equation}
where $\mathcal{F}^{-1}(\cdot)$ denotes the inverse Fourier transform.
In the case of multiple channels, the approach independently learns a single filter per channel. Since the support of the learned filter is constrained to be smaller than the learning region, the maximum response on the training region reflects the reliability of the learned filter~\cite{Lukezic_CVPR_2017}. 
These values are used as per-channel weights $w_d$ in (\ref{eq:cf_localization}) for improved target localization.

Note that the constrained learning~\cite{Lukezic_CVPR_2017} estimates a filter implicitly padded with zeros to match the learning region size. In contrast to the  standard approach to filter learning like e.g.,~\cite{henriques2015tracking} and multiplying with a mask post-hoc, the padding is explicitly enforced during learning, resulting in an increased filter robustness. 
We make an observation that adding or removing the zeros at filter borders keeps the filter unchanged, thus correlation on an arbitrary large region via FFT is possible by zero padding the filter to match the size.
These properties make the constrained learning an excellent candidate to train the short-term component (Section~\ref{sec:detector}) as well as the detector (Section~\ref{sec:detector}) in a long-term tracker. 
 
\subsection{The short-term component}\label{sec:short-term-tracker}

\begin{figure}[t]
\begin{center}
	\includegraphics[width=0.9\linewidth]{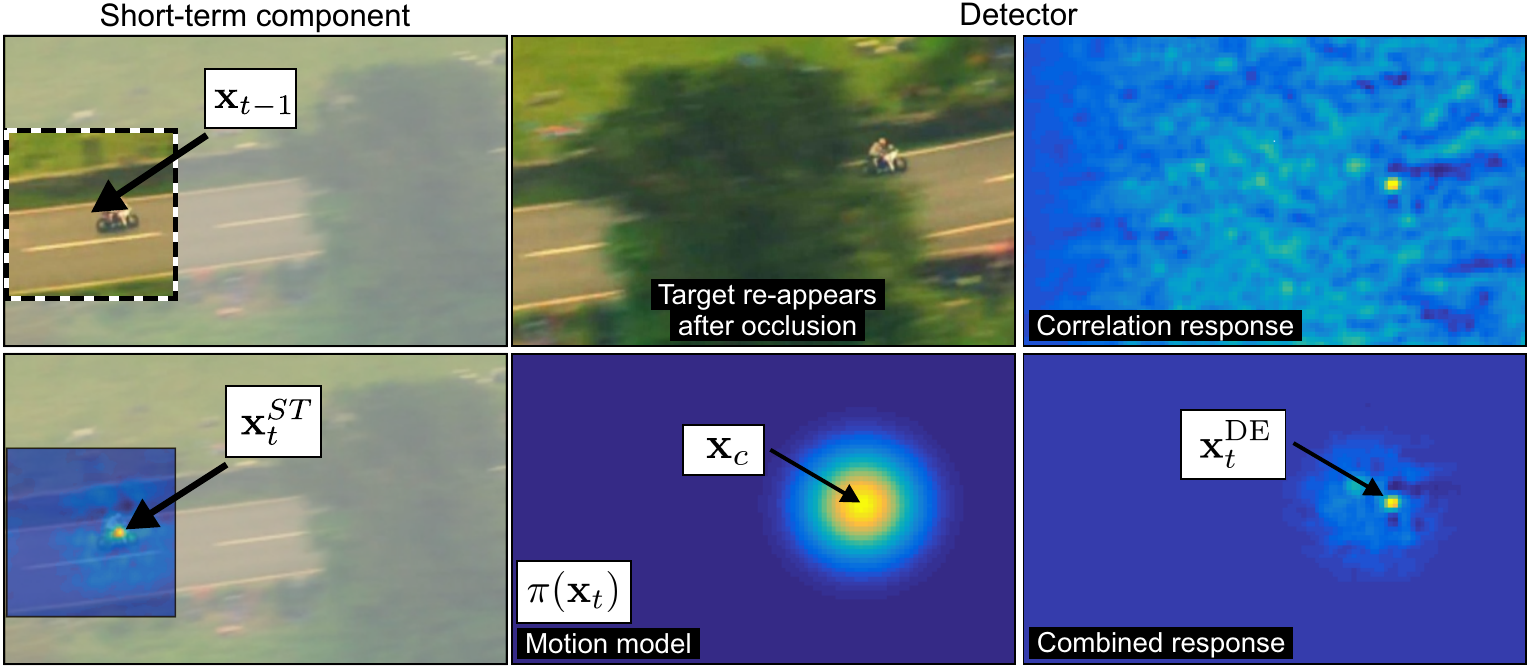}
\end{center}
   \caption{The short-term component (left) estimates the target location at the maximum response of its DCF within a search region centered at the estimate in the previous frame. The detector (right) estimates the target location as the maximum in the whole image of the response of its DCF multiplied by the motion model $\pi (\mathbf{x}_{t})$. If tracking fails, the prior $\pi (\mathbf{x}_{t})$ spreads.}
\label{fig:st_detector}
\end{figure}

The CSRDCF~\cite{Lukezic_CVPR_2017} tracker is used as the short-term component in FuCoLoT. The short-term component is run within a search region centered on the target position predicted from the previous frame. 
The new target position hypothesis $\mathbf{x}^\mathrm{ST}_t$ is estimated as the location of the maximum of the correlation response between the short-term filter $\mathbf{h}^\mathrm{ST}_{t}$ and the features extracted from the search region (see Figure~\ref{fig:st_detector}, left).
 
The visual model of the short-term component $\mathbf{h}^\mathrm{ST}$ is updated by a weighted running average
\begin{equation} \label{eq:filter-update}
\mathbf{h}^\mathrm{ST}_{t+1} = (1-\eta)\mathbf{h}^\mathrm{ST}_{t} + \eta \tilde{\mathbf{h}}^\mathrm{ST}_{t},
\end{equation}
where $\mathbf{h}^\mathrm{ST}_{t}$ is the correlation filter used to localize the target, $\tilde{\mathbf{h}}^\mathrm{ST}_{t}$ the filter estimated by constrained filter learning  (Section~\ref{sec:csrdcf}) in the current frame, and $\eta$ is the update factor.

\subsection{Tracking uncertainty estimation} \label{sec:localization_quality}

Tracking uncertainty estimation is crucial for minimizing short-term visual model contamination as well as for activating target re-detection after events like occlusion. We propose a self-adaptive approach for tracking uncertainty based on the maximum correlation response.

\begin{figure}[t]
\begin{center}
	\includegraphics[width=0.7\linewidth]{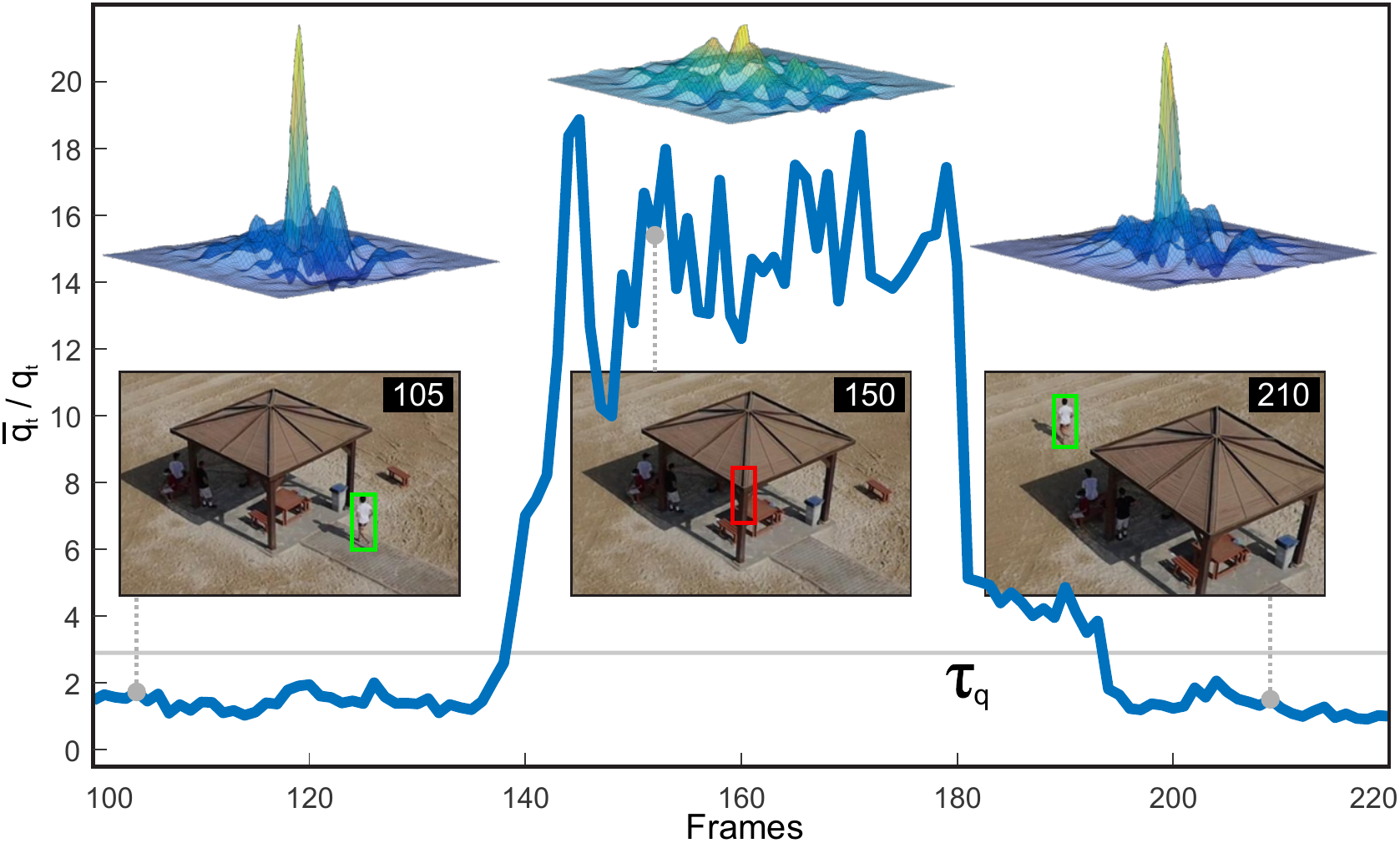}
\end{center}
   \caption{Localization uncertainty measure (\ref{eq:failure_detection_ratio}), reflects the correlation response peak strength relative to its past values. The measure rises fast during occlusion and drops immediately afterwards.}
\label{fig:failure-detection}
\end{figure}

Confident localization produces a well expressed local maximum in the correlation response $\mathbf{r}_t$, which can be measured by the peak-to-sidelobe ratio $\mathrm{PSR}(\mathbf{r}_t)$ \cite{bolme2010visual} and by the peak absolute value $\mathrm{max}(\mathbf{r}_t)$. 
Empirically, we observed that multiplying the two measures leads to improved performance, therefore the localization quality is defined as the product
\begin{equation} \label{eq:response_quality}
q_t = \mathrm{PSR}(\mathbf{r}_t) \cdot \mathrm{max}(\mathbf{r}_t).
\end{equation} 
The following observations were used in design of tracking uncertainty (or failure) detection:
(i) relative value of the localization quality $q_t$ depends on target appearance changes and is only a weak indicator of tracking uncertainty, and (ii) 
events like occlusion occur on a relatively short time-scale and are reflected in a significant reduction of the localization quality.
Let $\overline{q}_t$ be the average localization quality computed over the recent $N_{q}$ confidently tracked frames. Tracking is considered uncertain if the ratio between $\overline{q}_t$ and ${q}_t$ exceeds a predefined threshold $\tau_q$, i.e., 
\begin{equation} \label{eq:failure_detection_ratio}
\overline{q}_t/q_t > \tau_q.
\end{equation}
In practice, the ratio between the average and current localization quality significantly increases during occlusion, indicating a highly uncertain tracking, and does not require  threshold fine-tuning (an example is shown in Figure~\ref{fig:failure-detection}).

\subsection{Target loss recovery}\label{sec:detector}

A visual model not contaminated by false training examples is desirable for reliable re-detection after a long period of target loss. 
The only known certainly uncontaminated filter is the one learned at initialization. However, for a short-term occlusions, the most recent uncontaminated model would likely yield a better detection. 
While contamination of the short-term visual model (Section~\ref{sec:short-term-tracker}) is reduced by the long-term system (Section~\ref{sec:tracking_fclt}), it cannot be prevented. 
We thus maintain as set of several filters $\mathcal{H}^{\mathrm{DE}} = \{ \mathbf{h}_{i}^{\mathrm{DE}} \}_{i\in 1,\ldots N_{\mathrm{DE}}}$ 
updated at different \textit{temporal scales} to deal with potential model contamination.

The filters updated frequently learn recent appearance changes, while the less frequently updated filters increase robustness to learning during potentially undetected tracking failure.
In our approach, one of the filters is never updated (the initial filter), which guarantees full recovery from potential contamination of the updated filters if a view similar to the initial training image appears. The $i$-th filter is updated every $n_{i}^{\mathrm{DE}}$ frames similarly as the short-term filter:
\begin{equation}  \label{eq:detector_update}
\mathbf{h}^{\mathrm{DE}}_{i} = (1-\eta) \mathbf{h}^{\mathrm{DE}}_{i} + \eta \tilde{\mathbf{h}}^{\mathrm{ST}}_t.
\end{equation}
 
A random-walk motion model is added as a principled approach to modeling the growth of the target search region size. The target position prior $\pi(\mathbf{x}_t)=\mathcal{N}(\mathbf{x}_t; \mathbf{x}_{c}, \Sigma_t)$ at time-step $t$ is a Gaussian with a diagonal covariance $\Sigma_t=\mathrm{diag}(\sigma_{xt}^2, \sigma_{yt}^2)$ centered at the last confidently estimated position $\mathbf{x}_{c}$. 
The variances in the motion model gradually increase with the number of frames $\Delta_t$ since the last confident estimation, i.e., $[\sigma_{xt}, \sigma_{yt}] = [x_w, x_h] \alpha_{s}^{\Delta_t}$, where $\alpha_{s}$ is scale increase parameter, $x_w$ and $x_h$ are the target width and height, respectively.

During target re-detection, a filter is selected from $\mathcal{H}^{\mathrm{DE}}$ and correlated with features extracted from the entire image. 
The detected position $\mathbf{x}^\mathrm{DE}_t$ is estimated as the location maximum of the correlation response multiplied with the motion prior $\pi(\mathbf{x}_t)$ as shown in Figure~\ref{fig:st_detector} (right). 
For implementation efficiency only a single filter is evaluated on each image. The algorithm cycles through all filters in $\mathcal{H}^{\mathrm{DE}}$ and all target size scales $\mathcal{S}^{\mathrm{DE}}$ in subsequent images until the target is detected. 
In practice this means that all filters are evaluated approximately within a second of the sequence (Section~\ref{sec:implementation}).

\subsection{Tracking with FuCoLoT}  \label{sec:tracking_fclt}

The FuCoLoT integrates the short-term component (Section~\ref{sec:short-term-tracker}), the uncertainty estimator (Section~\ref{sec:localization_quality}) and target recovery (Section~\ref{sec:detector})  as follows.

\textbf{Initialization.} The long-term tracker is initialized in the first frame and the learned initialization model $\mathbf{h}^\mathrm{ST}_1$ is stored.
In the remaining frames, $N_\mathrm{DE}$ visual models are maintained at different time-scales for target localization and detection $\{ \mathbf{h}_{i}^{\mathrm{DE}} \}_{i\in 1,\ldots N_{\mathrm{DE}}}$, where the model updated at every frame is the short-term visual model, i.e., $\mathbf{h}^\mathrm{ST}_t=\mathbf{h}_{N_\mathrm{DE}}^\mathrm{DE}$, and the model that is never updated is equal to the initialization model, i.e., $\mathbf{h}_{1}^\mathrm{DE}=\mathbf{h}^\mathrm{ST}_1$.

\textbf{Localization.} A tracking iteration at frame $t$ starts with the target position $\mathbf{x}_{t-1}$ from the previous frame, a tracking quality score $q_{t-1}$ and the mean $\overline{q}_{t-1}$ over the recent $N_q$ confidently tracked frames. 
A region is extracted around $\mathbf{x}_{t-1}$ in the current image and the correlation response is computed using the short-term component model $\mathbf{h}^\mathrm{ST}_{t-1}$ (Section~\ref{sec:short-term-tracker}). 
Position $\mathbf{x}^\mathrm{ST}_t$ and localization quality $q^\mathrm{ST}_t$ (\ref{eq:response_quality}) are estimated from the correlation response $\mathbf{r}_{t}^{\mathrm{ST}}$. 
When tracking is confident at $t-1$, i.e., the uncertainty (\ref{eq:failure_detection_ratio}) $\overline{q}_t/q_t$ was smaller than $\tau_q$, only the short-term component is run. Otherwise the detector (Section~\ref{sec:detector}) is activated as well to address potential target loss. 
The detector filter $\mathbf{h}^\mathrm{DE}_i$ is chosen from the sequence of stored detectors $\mathcal{H}^{\mathrm{DE}}$ and correlated with the features extracted from the entire image. 
The detection hypothesis $\mathbf{x}^\mathrm{DE}_t$ is obtained as the location of the maximum of the correlation multiplied by the motion model $\pi(\mathbf{x}_t)$, while the localization quality $q^\mathrm{DE}_t$ (\ref{eq:response_quality}) is computed only on the correlation response. 

\textbf{Update.} In case the detector has not been activated, the short-term position is taken as the final target position estimate. 
Alternatively, both position hypotheses, i.e., the position estimated by the short-term component as well as the position estimated by the detector, are considered. The final target position is estimated as the one with higher quality score, i.e., 
\begin{equation}
 ( \mathbf{x}_t,q_t ) = \left\{ {\begin{array}{*{20}{c}}
				(\mathbf{x}^\mathrm{ST}_t,q^\mathrm{ST}_t) &;& q^\mathrm{ST}_t \geq q^\mathrm{DE}_t\\
				(\mathbf{x}^\mathrm{DE}_t,q^\mathrm{DE}_t) &;& \mathrm{otherwise}
\end{array}.} \right. 
\end{equation}
If the estimated position is reliable (\ref{eq:failure_detection_ratio}), a constrained filter $\tilde{\mathbf{h}}^\mathrm{ST}_{t}$ is estimated according to~\cite{Lukezic_CVPR_2017} and the short-term component (\ref{eq:filter-update}) and detector (\ref{eq:detector_update}) are updated. Otherwise the models are not updated, i.e., $\eta=0$ in (\ref{eq:filter-update}) and (\ref{eq:detector_update}).

\section{Experiments}  \label{sec:experiments}

\subsection{Implementation details}  \label{sec:implementation}

We use the same standard HOG~\cite{dalal_triggs_hog} and colornames~\cite{colornames_tip2009,danelljan2014adaptive} features in the short-term component and in the detector. 
All the parameters of the CSRDCF filter learning are the same as in~\cite{Lukezic_CVPR_2017}, including filter learning rate $\eta = 0.02$ and regularization $\lambda = 0.01$. 
{ We use 5 filters in the detector, updated with the following frequencies $\{ 0, 1/250, 1/50, 1/10, 1 \}$ and size scale factors $\{ 0.5, 0.7, 1, 1.2, 1.5, 2 \}$.
}

The random-walk motion model region growth parameter was set to $\alpha_{s}=1.05$. 
The uncertainty threshold was set to $\tau_{q}=2.7$ and the parameter ``recent frames'' was $N_{q}=100$. 
The parameters did not require fine tuning and were  kept  constant  throughout  all  experiments. 
Our Matlab implementation runs at 15~fps on OTB100~\cite{otb_pami2015}, 8~fps on VOT16~\cite{kristan_vot2016} and 6~fps on UAV20L~\cite{uav_benchmark_eccv2016} dataset. 
The experiments were conducted on an Intel Core i7 3.4GHz standard desktop.

\subsection{Evaluation on a long-term benchmark}  \label{sec:exp-long-term}

The long-term performance of the FuCoLoT is analyzed on the recent long-term benchmark UAV20L~\cite{uav_benchmark_eccv2016} that contains results of 11 trackers on 
20 long term sequences with average sequence length 2934 frames. To reduce clutter in the plots we include top-performing tracker 
SRDCF~\cite{srdcf_iccv2015} and all long-term trackers in the benchmark, i.e., MUSTER~\cite{muster_cvpr2015} and TLD~\cite{kalal_pami}. We add the most recent state-of-the-art long-term trackers CMT~\cite{CMT_CVPR2015}, LCT~\cite{LCT_CVPR2015}, and PTAV~\cite{ptav_iccv2017} in the analysis, as well as the recent state-of-the-art short-term DCF trackers CSRDCF~\cite{Lukezic_CVPR_2017}, CCOT~\cite{danelljan_eccv2016_ccot} and CNN-based MDNet~\cite{mdnet_cvpr2016} and SiamFC~\cite{siamfc_eccv16}.

\begin{figure}[t]
\begin{center}
	\includegraphics[width=.9\linewidth]{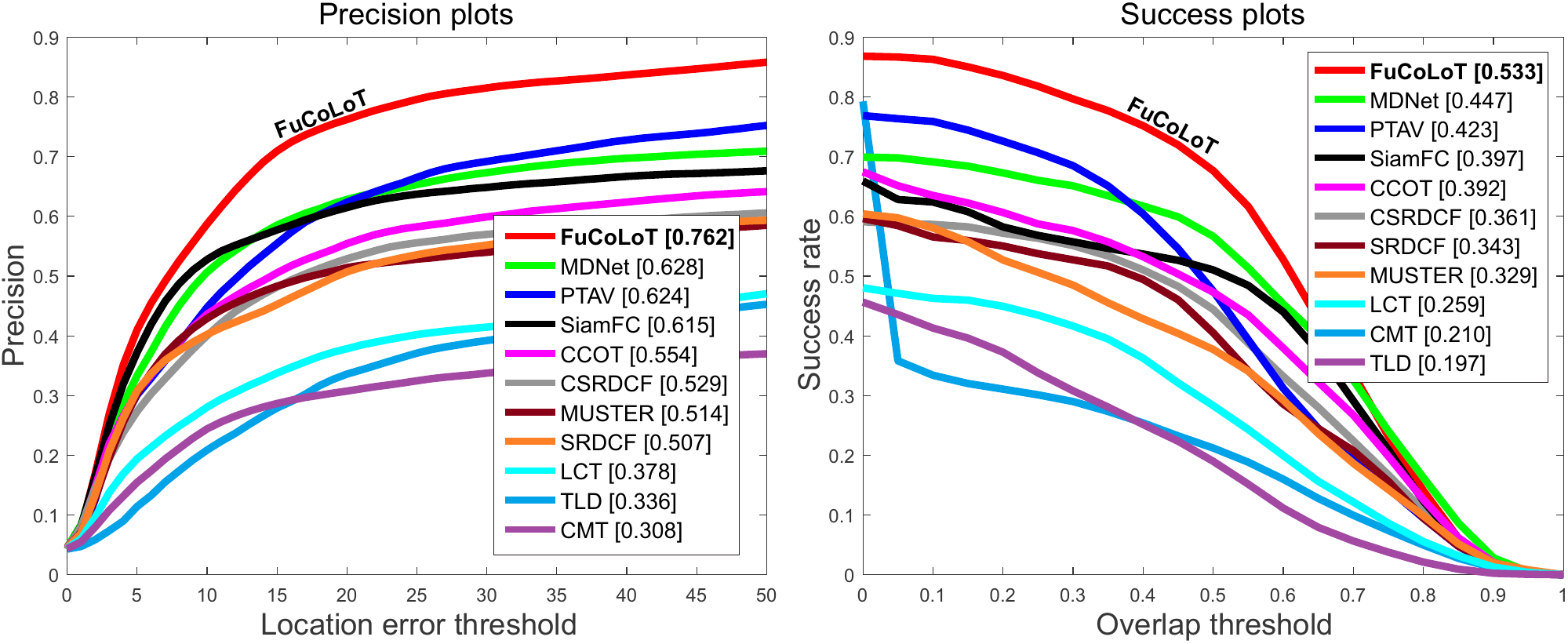}
\end{center}
   \caption{UAV20L~\cite{uav_benchmark_eccv2016} benchmark results. The precision plot (left) and the success plot (right).}
\label{fig:uav20L-graph}
\end{figure}

Results in Figure~\ref{fig:uav20L-graph} show that on benchmark, FuCoLoT by far outperforms  all top baseline trackers as well as all the recent long-term state-of-the-art. 
In particular FuCoLoT outperforms the recent long-term correlation filter LCT~\cite{LCT_CVPR2015} by $102\%$ in precision and $106\%$ in success measures. 
The FuCoLoT also outperforms the currently best-performing published long-term tracker PTAV~\cite{ptav_iccv2017} by over $22\%$ and $26\%$ in precision and success measures, respectively.
This is an excellent result especially considering that FuCoLoT does not apply deep features and backtracking like PTAV~\cite{ptav_iccv2017} and that it runs in near-realtime on a single thread CPU.
The FuCoLoT outperforms the second-best tracker on UAV20L, MDNet~\cite{mdnet_cvpr2016} which uses pre-trained network and runs at cca. 1fps, by $21\%$ in precision and $19\%$ in success measure.

Table~\ref{tab:uav20L-attributes} shows tracking performance in terms of the AUC measure for the twelve attributes annotated in the UAV20L benchmark. The FuCoLoT is the top performing tracker across all attributes,
including full occlusion and out-of-view, where it outperforms the second-best PTAV and MDNet by $29\%$ and $11\%$, respectively.
These attributes focus on the long-term tracker capabilities since they require target re-detection.

\begin{table*}[t]
\begin{center}
\scalebox{.68}{
\begin{tabular}{ l r r r r r r r r r r r r }
\hline
\multicolumn{1}{ p{0.077\linewidth}}{Tracker} & 
\multicolumn{1}{>{\centering}p{0.070\linewidth}}{Scale Var.} & 
\multicolumn{1}{>{\centering}p{0.070\linewidth}}{Aspect\\Change} & 
\multicolumn{1}{>{\centering}p{0.070\linewidth}}{Low\\Res.} & 
\multicolumn{1}{>{\centering}p{0.070\linewidth}}{Fast\\Motion} & 
\multicolumn{1}{>{\centering}p{0.070\linewidth}}{Full\\Occ.} & 
\multicolumn{1}{>{\centering}p{0.070\linewidth}}{Partial\\Occ.} & 
\multicolumn{1}{>{\centering}p{0.090\linewidth}}{Out-of-\\View} & 
\multicolumn{1}{>{\centering}p{0.070\linewidth}}{Back.\\Clutter} & 
\multicolumn{1}{>{\centering}p{0.070\linewidth}}{Illum.\\Var.} & 
\multicolumn{1}{>{\centering}p{0.070\linewidth}}{Viewp.\\Change} & 
\multicolumn{1}{>{\centering}p{0.070\linewidth}}{Camera\\Motion} & 
\multicolumn{1}{>{\centering}p{0.070\linewidth} }{Similar\\Object} \\ 
\hline
FuCoLoT & \first{0.526} & \first{0.500} & \first{0.472} & \first{0.400} & \first{0.460} & \first{0.526} & \first{0.488} & \first{0.538} & \first{0.513} & \first{0.499} & \first{0.528} & \first{0.573} \\
MDNet & \second{0.438} & \third{0.385} & \third{0.357} & \second{0.383} & 0.188 & \second{0.419} & \second{0.439} & 0.149 & \third{0.403} & \second{0.444} & \second{0.432} & \second{0.525} \\
PTAV & \third{0.416} & \second{0.410} & \second{0.390} & \third{0.349} & \second{0.357} & \third{0.415} & \third{0.389} & \second{0.435} & \second{0.430} & \third{0.418} & \third{0.420} & 0.426 \\
SiamFC & 0.383 & 0.328 & 0.242 & 0.264 & \third{0.237} & 0.364 & 0.386 & \third{0.239} & 0.371 & 0.356 & 0.383 & 0.436 \\
CCOT & 0.378 & 0.322 & 0.277 & 0.275 & 0.183 & 0.368 & 0.352 & 0.188 & 0.382 & 0.330 & 0.380 & \third{0.463} \\
CSRDCF & 0.346 & 0.293 & 0.232 & 0.194 & 0.210 & 0.339 & 0.326 & 0.227 & 0.359 & 0.308 & 0.348 & 0.403 \\
SRDCF & 0.332 & 0.270 & 0.228 & 0.197 & 0.170 & 0.320 & 0.329 & 0.156 & 0.295 & 0.303 & 0.327 & 0.397 \\
MUSTER & 0.314 & 0.275 & 0.278 & 0.206 & 0.200 & 0.305 & 0.309 & 0.230 & 0.242 & 0.318 & 0.307 & 0.342 \\
LCT & 0.244 & 0.201 & 0.183 & 0.112 & 0.151 & 0.244 & 0.249 & 0.156 & 0.232 & 0.225 & 0.245 & 0.283 \\
CMT & 0.208 & 0.169 & 0.139 & 0.199 & 0.134 & 0.173 & 0.184 & 0.104 & 0.146 & 0.212 & 0.187 & 0.203 \\
TLD & 0.193 & 0.196 & 0.159 & 0.235 & 0.154 & 0.201 & 0.212 & 0.111 & 0.167 & 0.188 & 0.202 & 0.225 \\
\hline
\end{tabular}
}
\end{center}
\caption{Tracking performance (AUC measure) for fourteen tracking attributes and eleven trackers on the UAV20L~\cite{uav_benchmark_eccv2016}.}
\label{tab:uav20L-attributes}
\end{table*}

\begin{figure*}[t]
\begin{center}
	\includegraphics[width=1\linewidth]{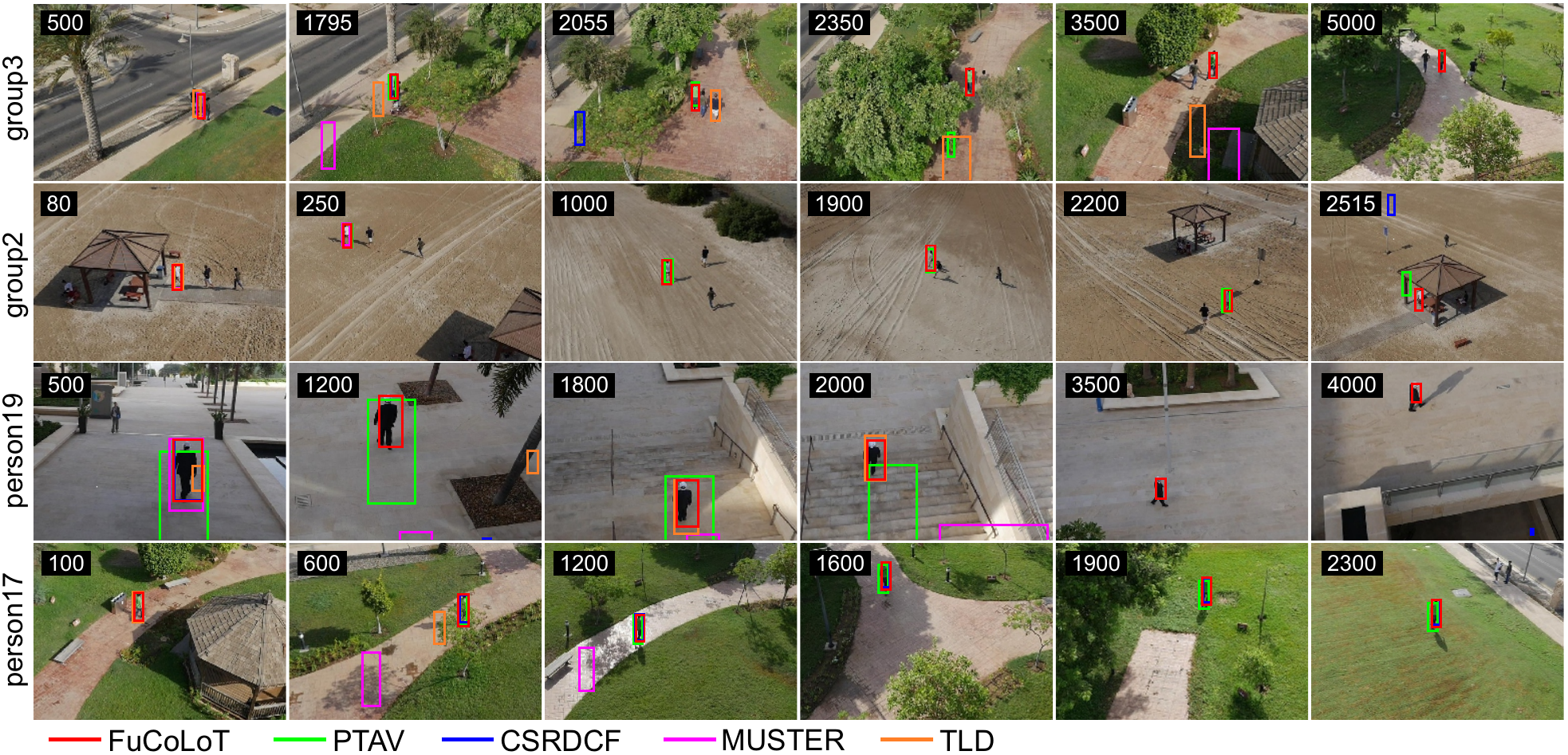}
\end{center}
   \caption{{ Qualitative results of the FuCoLoT and four state-of-the-art trackers on four sequences from~\cite{uav_benchmark_eccv2016}. The selected sequences contain full occlusions which highlight the re-detection ability of a tracker.}}
\label{fig:long_term_qualitative}
\end{figure*}

Figure~\ref{fig:long_term_qualitative} shows qualitative tracking examples for the  FuCoLoT and four state-of-the-art trackers: 
PTAV~\cite{ptav_iccv2017}, CSRDCF~\cite{Lukezic_CVPR_2017}, MUSTER~\cite{muster_cvpr2015} and TLD~\cite{kalal_pami}. 
In all these sequences the target becomes fully occluded at least once. FuCoLot is the only tracker that is able to successfully track the target throughout Group2, Group3 and Person19 sequences, which shows the strength of the proposed correlation filter based detector. In Person 17 sequence, the occlusion is shorter, thus the short-term CSRDCF~\cite{Lukezic_CVPR_2017} and long-term PTAV~\cite{ptav_iccv2017} are able to track as well.

\subsection{Re-detection capability experiment} \label{sec:redetection-experiment}

\begin{figure}[!ht]
\begin{center}
	\includegraphics[width=0.9\linewidth]{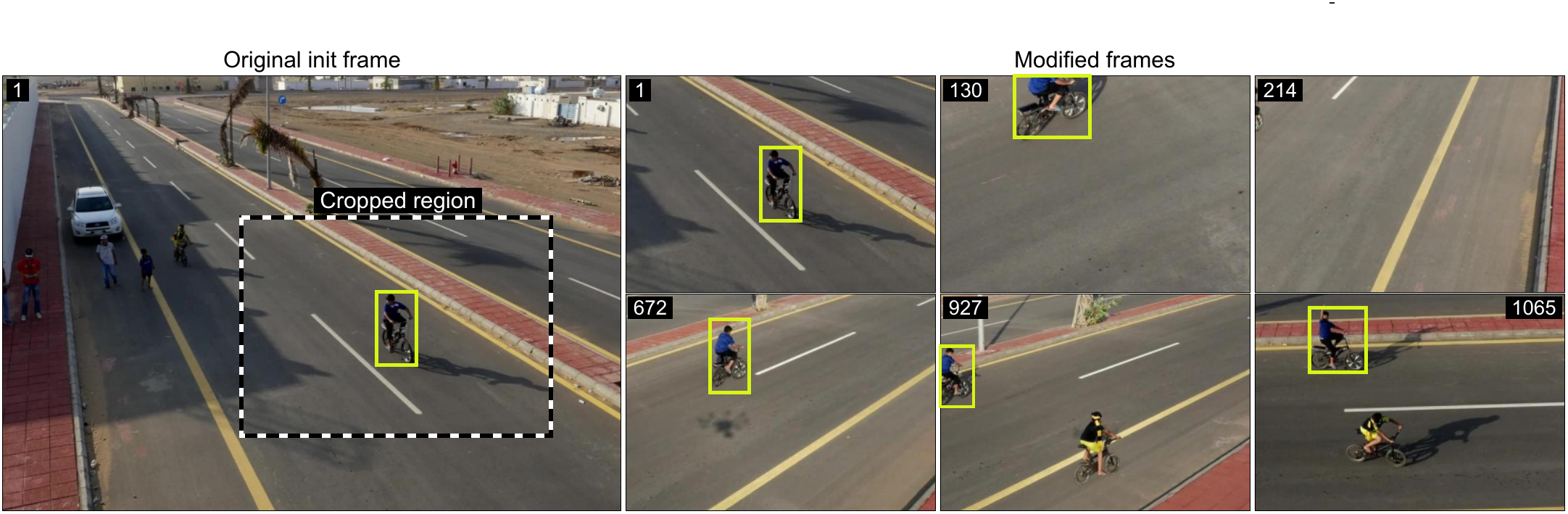}
\end{center}
   \caption{Re-detection experiment - an example of the modification of a sequence. Yellow bounding-boxes denote the ground-truth position of the target. The target leaves the field-of-view more frequently in the dataset with cropped images.}
\label{fig:reinit-example}
\end{figure}

In the original UAV20L~\cite{uav_benchmark_eccv2016} dataset, the target disappears and reappears only 39 times, resulting in only 4\% of frames with the target absent. This setup does not significantly expose the target re-detection capability of the tested trackers. To address this, we have cropped the images in all sequences to 40\% of their size around the center at target initial position. An example of the dataset modification is shown in Figure~\ref{fig:reinit-example}. This modification increased the target disappearance/reappearance to 114 cases, and the target is absent in 34\%  of the frames.

The top six trackers from Section~\ref{sec:exp-long-term} and a long-term baseline TLD~\cite{kalal_pami} were re-evaluated on the modified dataset (results in Table~\ref{tab:redetection_experiment}). 
The gap between the FuCoLoT and the other trackers is further increased. FuCoLoT outperforms the second-best MDNet~\cite{mdnet_cvpr2016} by 30\% and the recent long-term state-of-the-art tracker PTAV~\cite{ptav_iccv2017} by 47\%. 
Note that FuCoLoT outperforms all CNN-based trackers using only hand-crafted features, which speaks in favor of the highly efficient architecture.

\begin{table}[!ht]
\begin{center}
\begin{tabular}{c c c c c c c || c}
\hline
Tracker & FuCoLoT & MDNet & CCOT & PTAV & SiamFC & TLD & CSRDCF \\
AUC & {\bf 0.314} & 0.240 & 0.220 & 0.213 & 0.205 & 0.160 & 0.158 \\
\hline
\end{tabular}
\end{center}
\caption{Re-detection experiment on the UAV20L~\cite{uav_benchmark_eccv2016} dataset with images cropped to increase the number of times the target leaves and re-enters the field of view.} \label{tab:redetection_experiment}
\end{table}

\subsection{Ablation study}  \label{sec:ablation_study} 

Several modifications of the FuCoLoT detector were tested to expose the contributions of different parts in our architecture. Two variants used the filter extracted at initialization in the detector with a single scale detection ($\mathrm{FuCoLoT_{D^{0}S^{1}}}$) and multiple scale detection ($\mathrm{FuCoLoT_{D^{0}S^{M}}}$) and  one variant used the most recent filter from the short-term tracker in the detector with multiple scale detection ($\mathrm{FuCoLoT_{D^{ST}S^{M}}}$). The results are summarized in Table~\ref{tab:ablation-study}. 
  
In single-scale detection, the most recent short-term filter ($\mathrm{FuCoLoT_{D^{ST}S^{1}}}$) marginally improves the performance of the ($\mathrm{FuCoLoT_{D^{0}S^{1}}}$) and achieves $0.499$ AUC. The performance improves to  $0.505$ AUC by adding multiple scales search in the detector ($\mathrm{FuCoLoT_{D^{ST}S^{M}}}$) and further improves to $0.533$ AUC when considering filters with variable temporal updating in the detector ($\mathrm{FuCoLoT}$). For reference, all FuCoLoT variants significantly outperform the FuCoLOT short-term tracker without our detector, i.e. the CSRDCF~\cite{Lukezic_CVPR_2017} tracker.

\begin{table}[t]
\begin{center}
\begin{tabular}{c c c c c  || c}
\hline
Tracker & $\mathrm{FuCoLoT_{D^{0}S^{1}}}$ & $\mathrm{FuCoLoT_{D^{ST}S^{1}}}$ & $\mathrm{FuCoLoT_{D^{ST}S^{M}}}$ & $\mathrm{FuCoLoT}$ & CSRDCF~\cite{Lukezic_CVPR_2017} \\
AUC & 0.489 & 0.499 & 0.505 & {\bf 0.533}  & 0.361 \\
\hline
\end{tabular}
\end{center}
\caption{Ablation study of the FuCoLoT tracker on UAV20L~\cite{uav_benchmark_eccv2016}.} 
\label{tab:ablation-study}
\end{table}

\subsection{Performance on short-term benchmarks}  \label{sec:exp-short-term}

For completeness, we first evaluate the performance of FuCoLoT on the popular short-term benchmarks: OTB100~\cite{otb_pami2015}, and VOT2016~\cite{kristan_vot2016}. 
A standard no-reset evaluation (OPE~\cite{otb_pami2015}) is applied to focus on long-term behavior: a tracker is initialized in the first frame and left to track until the end of the sequence. 

\begin{figure}[t]
\begin{center}
	\includegraphics[width=.9\linewidth]{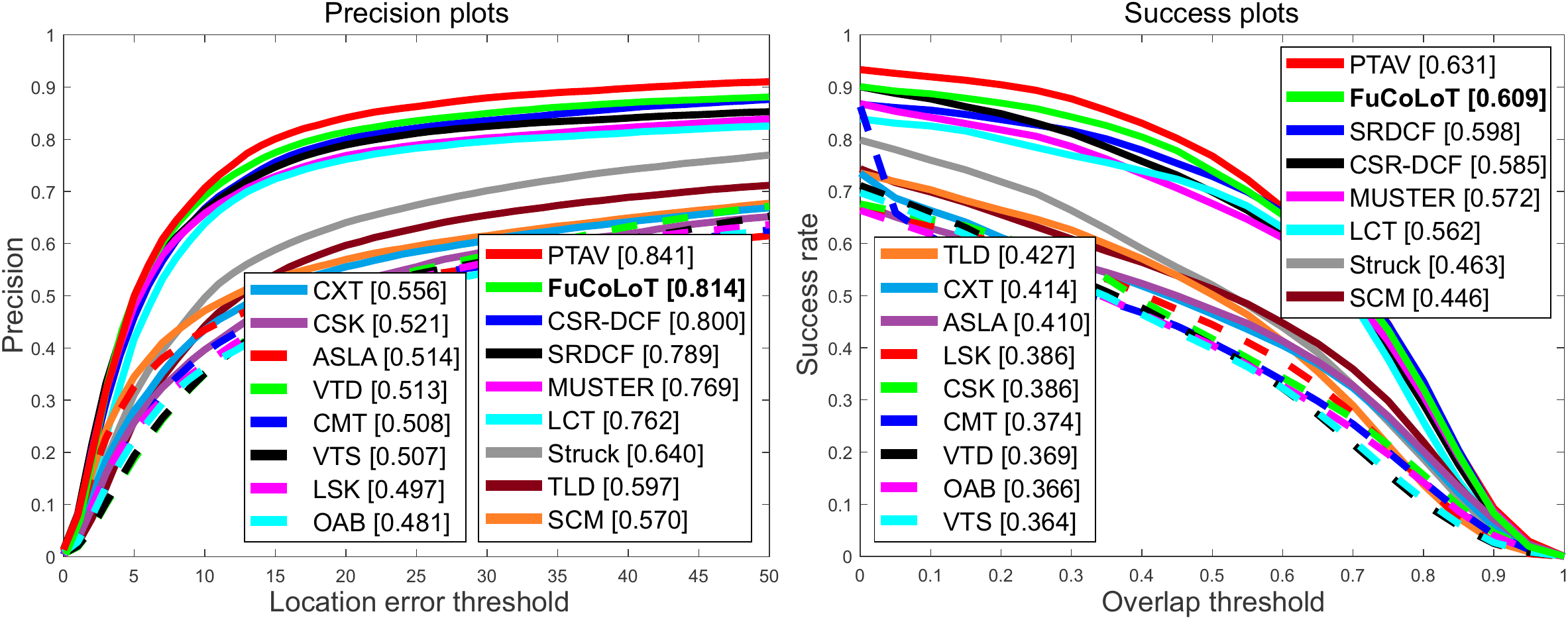}
\end{center}
   \caption{OTB100~\cite{otb_pami2015} benchmark results. The precision plot (left) and the success plot (right).}
\label{fig:otb-graph}
\end{figure}

\begin{figure}[t]
\begin{center}
	\includegraphics[width=.9\linewidth]{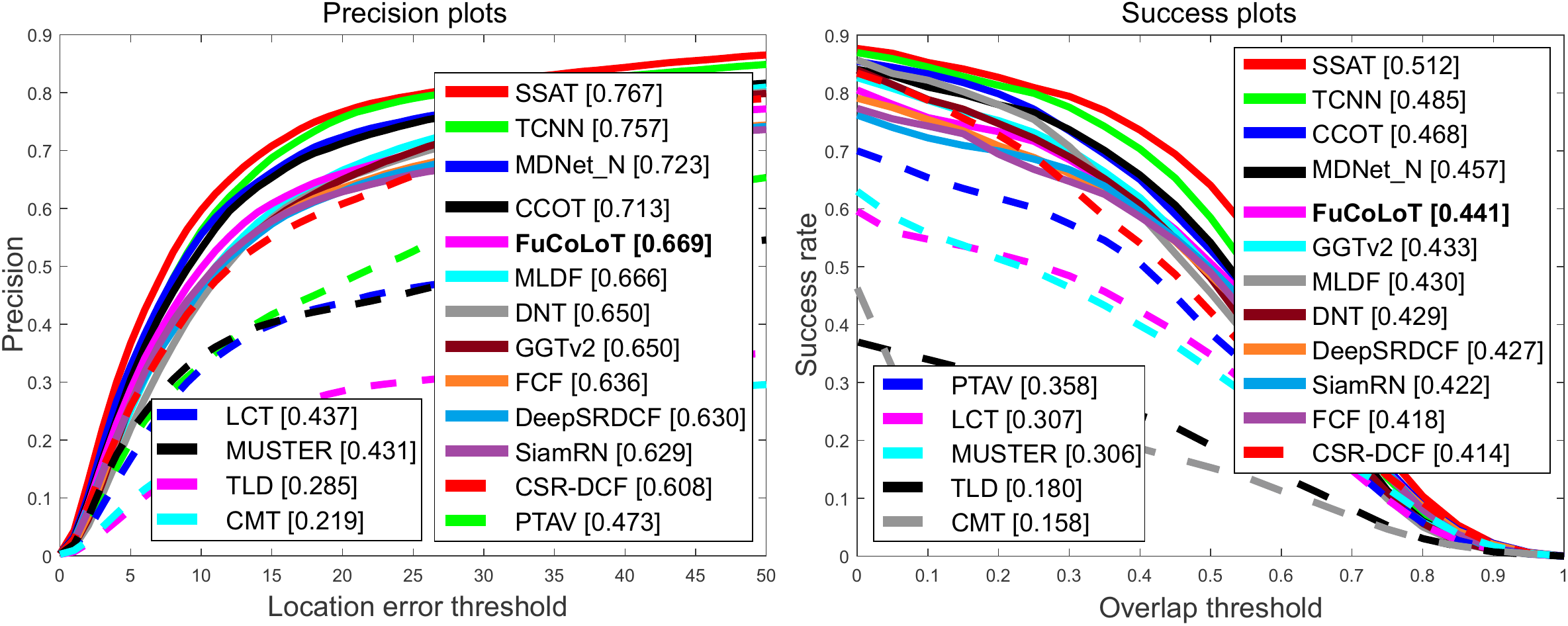}
\end{center}
   \caption{VOT2016~\cite{kristan_vot2016} benchmark results. The precision plot (left) and the success plot (right).}
\label{fig:vot-graph}
\end{figure}

Tracking  quality  is  measured  by  precision and  success  plots. 
The success plot shows all threshold values, the proportion  of  frames with the overlap between the predicted and ground truth bounding boxes as greater than a threshold. 
The results are summarized by areas under these plots which are shown in the legend.
The precision plots in Figures~\ref{fig:otb-graph} and~\ref{fig:vot-graph} show a similar statistics computed from the center error. 
The results in the legends are summarized by percentage of frames tracked with an center error less than 20 pixels.

The benchmarks results have some long-term trackers and the most recent PTAV~\cite{ptav_iccv2017} -- the currently best-performing published long-term tracker. 
Note that PTAV applies preemptive tracking with backtracking and requires future frames to predict position of the tracked object which limits its applicability.

{\bf OTB100~\cite{otb_pami2015}} contains results of 29 trackers evaluated on 100 sequences with average sequence length of 589 frames. 
We show only the results for top-performing recent baselines, and recent top-performing state-of-the-art trackers 
SRDCF~\cite{srdcf_iccv2015}, MUSTER~\cite{muster_cvpr2015}, LCT~\cite{LCT_CVPR2015} PTAV~\cite{ptav_iccv2017} and CSRDCF~\cite{Lukezic_CVPR_2017}. 

The FuCoLoT ranks among the top on this benchmark (Figure~\ref{fig:otb-graph}) outperforming all baselines as well as state-of-the-art SRDCF, CSRDCF and MUSTER. 
Using only handcrafted features, the FuCoLoT achieves comparable performance to the PTAV~\cite{ptav_iccv2017} which uses deep features for re-detection and backtracking.

{\bf VOT2016~\cite{kristan_vot2016}} is a challenging recent short-term tracking benchmark which evaluates 70 trackers on 60 sequences with the average sequence length of 358 frames. 
The dataset was created using a methodology that selected sequences which are difficult to track, thus the target appearance varies much more than in other benchmarks. 
In the interest of visibility, we show only top-performing trackers on no-reset evaluation, i.e.,
SSAT~\cite{mdnet_cvpr2016,kristan_vot2016}, TCNN~\cite{mdnet_cvpr2016,kristan_vot2016}, CCOT~\cite{danelljan_eccv2016_ccot}, MDNetN~\cite{mdnet_cvpr2016,kristan_vot2016}, GGTv2~\cite{GGT_tcyb2017}, MLDF~\cite{wang_iccv2015}, DNT~\cite{dnt_tip2017}, DeepSRDCF~\cite{srdcf_iccv2015}, SiamRN~\cite{siamfc_eccv16} and FCF~\cite{kristan_vot2016}.
We add CSRDCF~\cite{Lukezic_CVPR_2017} and the long-term trackers TLD~\cite{kalal_pami}, LCT~\cite{LCT_CVPR2015}, MUSTER~\cite{muster_cvpr2015}, CMT~\cite{CMT_CVPR2015} and PTAV~\cite{ptav_iccv2017}. 

The FuCoLoT is ranked fifth (Figure~\ref{fig:vot-graph}) according to the tracking success measure, outperforming 65 trackers, including trackers with deep features, CSR-DCF and PTAV.
Note that four trackers with better performance than FuCoLoT (SSAT, TCNN, CCOT and MDNetN) are computationally very expensive CNN-based trackers. 
They are optimized for accurate tracking on short sequences, without an ability for re-detection. 
The FuCoLoT outperforms all long-term trackers on this benchmark (TLD, CMT, LCT, MUSTER and PTAV).

\section{Conclusion}  \label{sec:conclusion}

A fully-correlational long-term tracker -- FuCoLot -- was proposed. FuCoLoT is the first long-term tracker that exploits the novel DCF constrained filter learning method \cite{Lukezic_CVPR_2017}. 
The constrained filter learning based detector is able to re-detect the target in the whole image efficiently. 
FuCoLoT maintains several correlation filters trained on different time scales that act as the detector components.
A novel mechanism based on the correlation response quality is used for tracking uncertainty estimation which drives interaction between the short-term component and the detector.

On  the UAV20L long-term benchmark~\cite{uav_benchmark_eccv2016} FuCoLoT outperforms the best method by over 19\%. 
Experimental evaluation on short-term  benchmarks~\cite{otb_pami2015,kristan_vot2016} showed state-of-the-art performance. The Matlab implementation running at 15 fps will be made publicly available.

{\footnotesize
\noindent {\bf Acknowledgement} 
This work was partly supported by the following research programs and projects: 
Slovenian research agency research programs and projects P2-0214 and J2-8175. 
Jiří Matas and Tomáš Vojíř were supported by The Czech Science Foundation Project GACR P103/12/G084 and Toyota Motor Europe.
}

\bibliographystyle{splncs04}
\bibliography{bib}

\end{document}